\pdfoutput=1
\documentclass[letterpaper, 10 pt, conference]{ieeeconf}  % Comment this line out if you need a4paper

\IEEEoverridecommandlockouts                              % This command is only needed if 
\usepackage{xcolor}
\usepackage{amsmath}
\usepackage{graphicx}
\usepackage{footnote}
\usepackage{algorithm}
\usepackage{algpseudocode}
\usepackage{romannum}
\usepackage{subcaption}
\DeclareFontFamily{OT1}{pzc}{}
\DeclareFontShape{OT1}{pzc}{m}{it}{<-> s * [1.10] pzcmi7t}{}
\DeclareMathAlphabet{\mathpzc}{OT1}{pzc}{m}{it}

\graphicspath{{./figs/}}

\title{\LARGE \bf
Learning a Skill-sequence-dependent Policy for Long-horizon Manipulation Tasks
}

\author{Zhihao Li\textsuperscript{1†}, Zhenglong Sun\textsuperscript{2†}, Jionglong SU\textsuperscript{3}, Jiaming Zhang\textsuperscript{1,4}% <-this % stops a space
\thanks{†Joint first authors}
\thanks{*This work was supported by the Shenzhen Science and Technology Innovation Commission (Grants No. JCYJ20170410172100520 and JCYJ20170818104502599). This work was also supported by the Shenzhen Institute of Artificial Intelligence and Robotics for Society (Grants No. 2019-INT020).}% <-this % stops a space
\thanks{$^{1}Z.\ Li$ and $J.\ Zhang$ is with Institute of Robotics and Intelligent Manufacturing, The Chinese University of Hong Kong (Shenzhen), Shenzhen, China {\tt\small {lizhihao, zhangjiaming}@cuhk.edu.cn}}%
\thanks{$^{2}Z.\ Sun$ is with School of Science and Engineering, The Chinese University of Hong (Shenzhen), Shenzhen, China  {\tt\small sunzhenglong@cuhk.edu.cn}}
\thanks{$^{3}J.\ Su$ is with School of AI and Advanced Computing, XJTLU Entrepreneur College (Taicang), Xi'an Jiaotong-Liverpool University, Suzhou, China {\tt\small Jionglong.Su@xjtlu.edu.cn}}
\thanks{$^{4}J.\ Zhang$ is with Shenzhen Institute of Artificial Intelligence and Robotics for Society, Shenzhen, China {\tt\small zhangjiaming@cuhk.edu.cn} }
}

\begin{document}

\maketitle
\begin{abstract}
In recent years, the robotics community has made substantial progress in robotic manipulation using deep reinforcement learning (RL). Effectively learning of long-horizon tasks remains a challenging topic. Typical RL-based methods approximate long-horizon tasks as Markov decision processes and only consider current observation (images or other sensor information) as input state. However, such approximation ignores the fact that skill-sequence also plays a crucial role in long-horizon tasks. In this paper, we take both the observation and skill sequences into account and propose a skill-sequence-dependent hierarchical policy for solving a typical long-horizon task. The proposed policy consists of a high-level skill policy (utilizing skill sequences) and a low-level parameter policy (responding to observation) with corresponding training methods, which makes the learning much more sample-efficient. Experiments in simulation demonstrate that our approach successfully solves a long-horizon task and is significantly faster than Proximal Policy Optimization (PPO) and the task schema methods.
\end{abstract}


\section{INTRODUCTION}
\par Due to advances in learning-based methods, robots are getting more capable of performing intricate tasks. In recent years, there is much progress in robot manipulation \cite{akkaya2019solving}, \cite{mahler2019learning}, \cite{yan2020learning}. However, long-horizon manipulation tasks remain an open challenge in robotics \cite{gupta2019relay}, \cite{pirk2020modeling}. Since positive samples are sparse, the policy always needs to spend much time exploring. As a result, the application of learning-based methods to long-horizon tasks is limited.
\begin{figure}[t] %H为当前位置，!htb为忽略美学标准，htbp为浮动图形
\centering %图片居中
\includegraphics[width=0.48\textwidth]{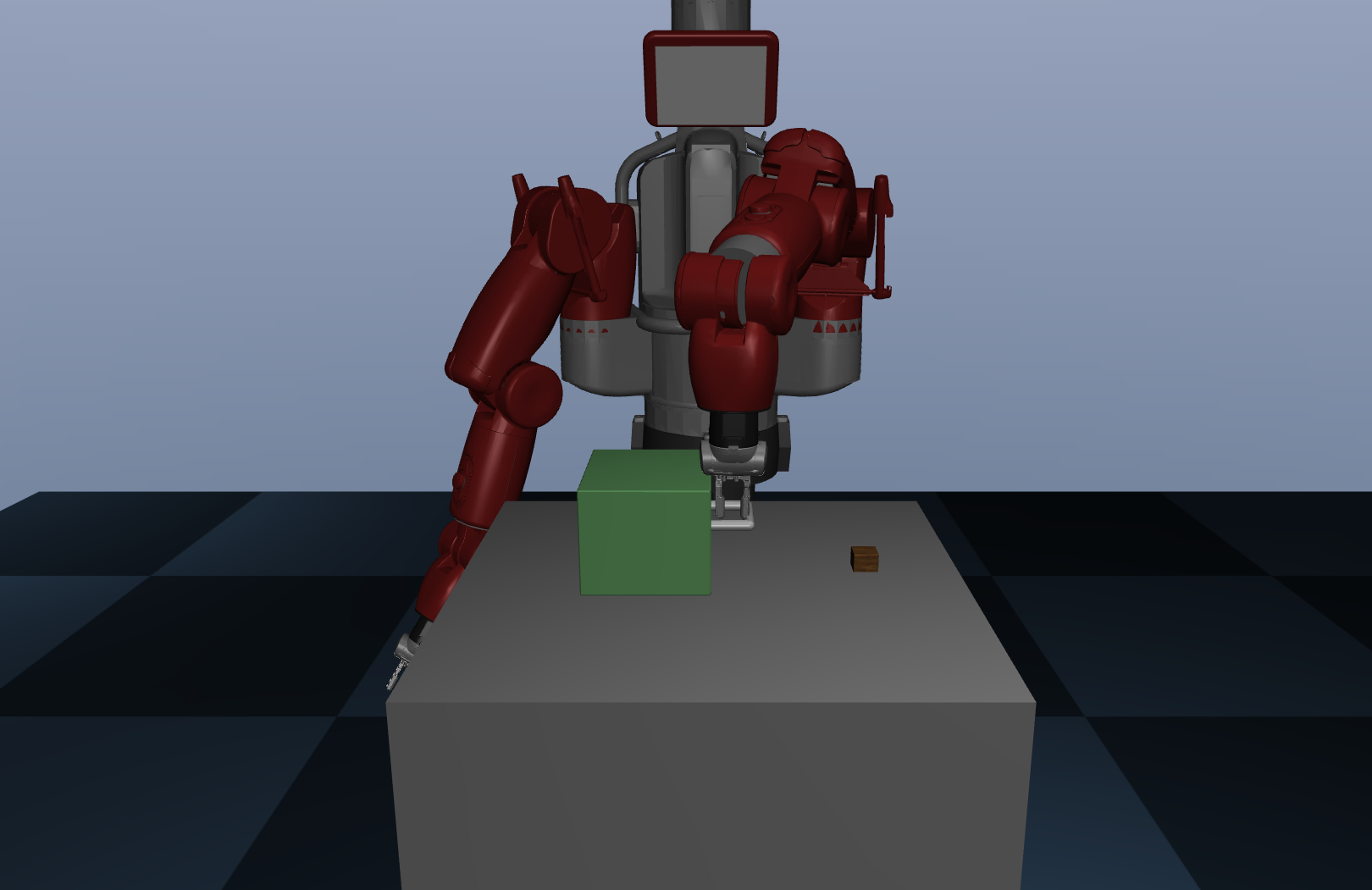} %插入图片，[]中设置图片大小，{}中是图片文件名
\caption{The long-horizon task in our experiment. We create the environment in Mujoco. The robot is Baxter. The long-horizon task is putting the block into the drawer and closing the drawer.} %最终文档中希望显示的图片标题
\label{Fig1} %用于文内引用的标签
\end{figure}
\par In robotics, a long-horizon task can be divided into meta-tasks, and each meta-task has a corresponding meta-skill, such as grasp, push, and put. Most methods use hierarchical policy for long-horizon tasks. The high-level policy determines the next skill, and the low-level policy chooses parameters for the skill. Most of these methods use observation or the current action as the input for the policy \cite{mandlekar2020learning}, \cite{chitnis2020efficient}, \cite{xia2020relmogen}, \cite{li2020learning}. There are also some works that consider future actions \cite{nair2019hierarchical}. An effective feature can greatly improve learning efficiency. Our insight is that skill sequences are crucial in long-horizon tasks, especially for the high-level policy. For example, people put block into a drawer. He needs to \textit{open} the drawer, \textit{grasp} the block, \textit{put} it into the drawer, and finally \textit{close} the drawer. People can not put the block into the drawer first without opening the drawer.
\par In view of this, we propose a modified hierarchical policy that considers both observation and skill sequences. The high-level policy generates the next skill according to historical skill sequences. The low-level parameter policy chooses parameters for the skill responding to observation. Pirk et al. also shared the same train of thought [5], but their algorithm is adapted for learning from demonstrations. Our method can be used in self-supervised learning.
\par The parameter policy is widely used in robotic, especially in robot grasp. A grasp is commonly parameterized by the grasping point on the object and the wrist orientation of the robotic hand \cite{ekvall2007learning}, \cite{morales2006integrated}. Dex-net2.0 \cite{Dex-Net} uses a center relative to the camera and an angle in the table plane to represent a grasp, and it samples grasp candidates and rank them using a neural network. GG-CNN \cite{GG} outputs a grasp configuration and a quality estimate for each pixel in the image using a small fully convolutional architecture. Redmon et al. \cite{redmon2015real} propose a neural network that can predict an oriented grasp rectangle. This policy can be extended to other contact-rich manipulation tasks. Bogdanovic et al. \cite{bogdanovic2020learning} propose a variable impedance control policy, whose outputs corresponds to the parameters of the individual controller. Most of them use neural networks as the parameter policy, it is also used in our policy. This process can be regarded as a supervised learning problem. The agent's exploration collects the data for this process, and there is no need for human intervention.
\par Choosing skills is a sequential decision making problem whose solution can be obtained using reinforcement learning(RL). It has impressive successes in game playing \cite{mnih2015human}, \cite{van2015deep}. The agent will choose the action according to the state for maximizing the reward in one episode. It was also used in learning dexterous manipulation. Kumar et al. \cite{kumar2016optimal} control a pneumatically-actuated tendon-driven 24-DoF hand by model-based reinforcement learning. RL can be trained to carry out many challenging tasks, such as valve rotation, box flipping, and door opening [20]. Therefore, the high-level skill policy uses the RL method.
\par However, it is noticeable two common problems may occur when using such hierarchical policy for learning. First, imbalanced data between the current action and next action reduces the stability of the training process. As historical data is used for skill sequence decision, there would be much more data for the first meta-task than the late meta-task. Here we address this problem by using under-sampling. Second, the inefficient exploration results in a lot of time spent in training. The traditional method(the high-level policy and the low-level policy explore together in one episode) has a bad performance in the long-horizon task. We propose a new exploration method making the low-level policy and the high-level policy explore separately. It can prevent the two policies from interfering with each other during exploration.
\par The evaluation of our approach is performed in a simulated environment consisting of a Baxter robot that needs to put a block into a drawer (Figure \ref{Fig1}). The result shows that the skill-sequence-dependent policy is suitable for long-horizon tasks. Compared to the other baselines, the method has a significantly higher success rate.

In summary, the contributions of this paper are as follows:
\begin{itemize}
    \item {A new policy is proposed, which takes into consideration both skill sequences and state.}
    \item{An improved exploration method for the framework is proposed. Moreover, the unbalanced data distribution problem is solved in exploration and exploitation.}
\end{itemize}

\section{RELATED WORK}
\par \textbf{Long-horizon task learning} Long-horizon task learning is popular in recent years. Krishnan et al. \cite{krishnan2016hirl} use hierarchical inverse reinforcement learning to solve the long-horizon tasks with delayed rewards. Nair et al. \cite{nair2019hierarchical} divide the task into subgoals and perform some robot manipulation tasks like push a block then open the door. Xia et al. \cite{xia2020relmogen} use a subgoal Generation Network to generate subgoal and a motion generator to execute low-level actions. However, they do not consider action sequences. Gupta et al. \cite{gupta2019relay} carry out the long-horizon robot manipulation tasks in the kitchen via reinforcement learning. Xiong et al. \cite{xiong2016robot} use and-or graph to perform a cloth-folding task. Edmonds et al. \cite{edmonds2019tale} completes an open-bottle task using the and-or graph. Pirk et al. \cite{pirk2020modeling} propose a deep learning network that learns dependencies and transitions across subtasks as action symbols from a set of demonstration videos. Our task is similar to their demonstrated task. However, the above methods focus on imitation learning and require human demonstrations. In contrast, our policy can learn to finish the task by itself. 

\par \textbf{Hierarchical policy} Hierarchical policy using temporally extended actions to reduce the sample complexity \cite{asada1996purposive}, \cite{hauskrecht2013hierarchical}. Yong et al. \cite{yang2018hierarchical} propose a hierarchical deep reinforcement learning algorithm, and they evaluate their algorithm using Pioneer 3AT robot in three different navigation tasks. Li et al. \cite{li2020hrl4in} propose a hierarchical RL architecture for interactive navigation tasks. Jain et al. \cite{jain2019hierarchical} use a high-level policy to decide the duration for which the low-level is executed and a low-level policy to output actions to execute. Chitnis et al. \cite{chitnis2020efficient} decompose tasks into learning a sequence of skills to execute and a policy to choose the parameterizations of the skills in a state-dependent manner. However, the input of their high-level policy is the task category, which does not offer flexibility enough. In contrast, our method chooses skills based on a historical skill sequence. The above methods define the low-level policy based on the reinforcement learning method. Some methods define low-level policy by the model-based method. Li et al. \cite{li2020learning} use a hierarchical control method for in-hand manipulation. They control 3-fingered hand by a two-level policy. Their high-level policy is based on reinforcement learning, but their low-level policy is a model-based method. They perform their method on the task of moving the object to the desired poses. Instead of reinforcement learning and model-based methods, our method regards the training problem of low-level policy as a supervised learning problem.

\section{METHODS}
%%%%%%%%%%%%%%%%%%%%%%%%%%%%%%%%%%%%%%%%%%
\par We focus on the problem of solving long-horizon robot manipulation tasks by learning-based methods. Long-horizon tasks have depended on both the skill sequences and the state, and the historical action sequences are more crucial for high-level policy. Therefore, they are used as the input for the high-level policy, which can determine the next skill, and the state is the input for the low-level policy, which is used to calculate parameters for the skill. This method significantly improves the speed of training and the success rate.
\par In this section, the details of our method are provided. First, the problem is formulated. Then we introduce the hierarchical policy that depends on skill sequences. After that, the improved exploration method and the approach for solving the imbalanced data distribution problem are provided. Moreover, full pseudocode is given.

\subsection{Problem Formulation}
 \par Humans always divide task into some sub-tasks when it is long-horizon and complex. The problem is formulated based on this fact.
\\ \textbf{Meta-tasks}: Long-horizon tasks can be divided into many meta-tasks. The meta-tasks are the primary tasks in manipulation, such as pushing something to somewhere and grasping something. $T_i$ denotes the meta-task.
\\ \textbf{Action}: The action for long-horizon tasks can be divided into two levels: skill and parameters for the skill. So it can be denoted by a tuple $(x_i, \theta_i)$. The $ x_i \in X $ is a meta-skill, and $\theta_i$ is the parameters for $ x_i $, $i$ denotes the $i$th skill in the skill set $X$. The skill is defined based on meta-tasks. It has a fixed action model, and the robot can take action by the skill with parameters. For example, the parameter for the top-grasp skill, the movement of the robot can be determined by the grasping position in the Cartesian coordinate system.
\\ \textbf{Input}: People determine what they do base on their historical skill sequences and determine how they do base on the states of object. So historical skill sequences are applied to choose the skill, and object states are used to calculate parameters. Historical skill sequences are denoted by $ h\_x(n,t) $, which means $n$ skills before time $t$, and it can be also denoted by $ (x_{t-1},x_{t-2},x_{t-3}...x_{t-n}) $. $o$ is object state. 
\\ \textbf{Reward}: The agent will receive rewards when the meta tasks finished. Different meta-tasks have different rewards. Reward at time $ t $ is denoted by $ r(t) $.

\begin{figure}[t] %H为当前位置，!htb为忽略美学标准，htbp为浮动图形
\centering %图片居中
\includegraphics[width=0.48\textwidth]{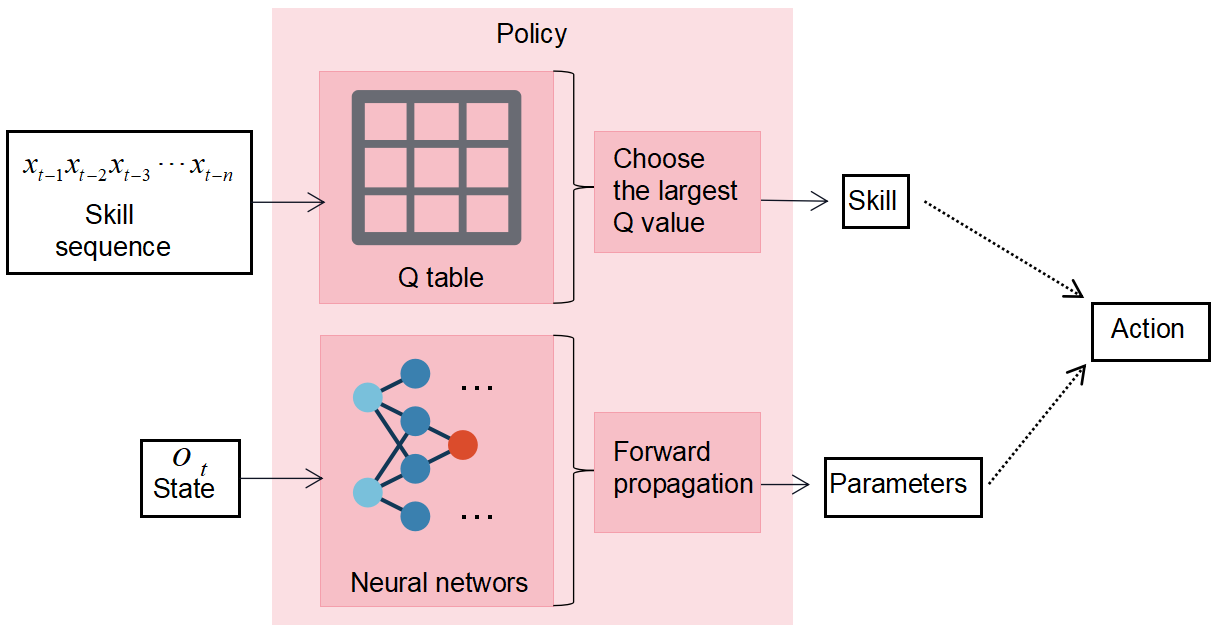}%插入图片，[]中设置图片大小，{}中是图片文件名
\caption{Our high-level policy is learned through the Q learning method. It can generate a Q table by training. The agent chooses the skill that has the largest Q value to execute. The low-level policy determines parameters for the skill through neural networks, and the state $o_t$ is the input for the low-level policy. The skill and parameters make up an action. The action can control Robots to finish long-horizon tasks.} %最终文档中希望显示的图片标题
\label{Fig2} %用于文内引用的标签
\end{figure}

\subsection{Hierarchical Policy}
\par Our Hierarchical policy has two levels. The high-level policy can choose a skill based on historical skill sequences. The low-level policy can determine the parameters for the skill. The overall of our policy is shown in Figure \ref{Fig2}.
\par High-level policy's input is the historical skill sequences  $ h\_x(n,t) $, which is finite and discrete. Furthermore, learning high-level policy is regarded as a sequential decision problem in our method. Q learning is a good solution for the sequential decision problem with a finite and discrete state\cite{melo2001convergence}. Therefore we use it to solve this problem. The agent will receive reward $ r(t) $ when the meta-task $ Ti $ is finished. The update rule for learning is:
\begin{equation}%加*表示不对公式编号
\begin{split}
Q_{t+1}(h\_x(n,t),x_i)=Q_t(h\_x(n,t),x_i)+lr(r(t)+\\
\gamma\max_{b\in X} Q_t(h\_x(n,t+1) ,b)-Q_t(h\_x(n,t),x_i))
\end{split}
\end{equation}
\par  where $Q_t$ denotes the Q value at time $t$, $b$ is the skill in skill sets, $lr$ is the learning rate in training, $\gamma$ denotes discount factor. With enough exploration, Q learning can converge and learn the optimal Q-values.
\par The low-level policy's input is the state of the object, which can be image or object position. Neural networks is used to obtain the parameters. If the agent gets a reward $ r(t) $, it will judge which meta-task is finished and save parameters $\theta_i$ and state $o$. The data is denoted as $(T_i,\theta_i,o)$. Parameters are the output of the network, and the state is the input of the network. $net(o)$ denote the network of policy. The cost function is: 
\begin{equation}%加*表示不对公式编号
\begin{split}
J=(net(o) - \theta_i)^2
\end{split}
\end{equation}
where $J$ denotes the cost of the low-level policy. $\theta_i$ denotes the skills' parameters that are collected by exploration. $net$ is the neural network. We use the gradient descent method to minimize the cost function.We give details on our policy in section \uppercase\expandafter{\romannumeral 4}.

\algnewcommand{\Object}[1]{%
  \State \textbf{Object:}
  \Statex \hspace*{\algorithmicindent}\parbox[t]{.8\linewidth}{\raggedright #1}
}
\algnewcommand{\Initialize}[1]{%
  \State \textbf{Initialize:}
  \Statex \hspace*{\algorithmicindent}\parbox[t]{.9\linewidth}{\raggedright #1}
}

\begin{algorithm}[t]
  \caption{The first exploration method}
  \label{Algorithm 1}
  \begin{algorithmic}[1]
    \Object{ low-level policy $\pi_l$, high-level policy $\pi_h$}
    \Initialize{ the environment $env$, $\epsilon$-greedy low-level policy $\epsilon$-$\pi_l$, $\epsilon$-greedy high-level policy $\epsilon$-$\pi_h$ }
    \While{$episode\_num< max\_episode\_num$}
        \State reset $h\_x$, $o$, $env$
        \State $episode\_num \leftarrow episode\_num+1$
        \While{not $done$}
            \State $\theta_i \leftarrow$ $\epsilon$-$ \pi_l (o)$ $x_i \leftarrow$ $\epsilon$-$\pi_h(h\_x)$
            \State $action \leftarrow (x_i,\theta_i)$
            \State  $r(t),o,done \leftarrow  env.step(action)$
            \If{$reward>0$}
                \State save $(T_i,o,\theta_i)$ 
            \EndIf
            \If{$done$}
                \State  $data \leftarrow under\_sample(data)$
                \State  Update $\pi_l $, $\pi_h$
            \EndIf
        \EndWhile 
    \EndWhile  
  \end{algorithmic}
\end{algorithm}

\begin{algorithm}[t]
  \caption{The improved exploration method}
  \label{Algorithm 2}
  \begin{algorithmic}[1]
    \Object{ low-level policy $\pi_l$, high-level policy $\pi_h$}
    \Initialize{ the environment $env$, $\epsilon$-greedy low-level policy $\epsilon$-$\pi_l$, $\epsilon$-greedy high-level policy $\epsilon$-$\pi_h$ }
    \While{$episode\_num< max\_episode\_num$}
        \State reset $h\_x$, $o$, $env$
        \State $flag \leftarrow random(0,1)$
        \State $episode\_num \leftarrow episode\_num+1$
        \While{not $done$}
            \If{$flag<0.5$}
                \State $\theta_i \leftarrow$ $\epsilon$-$ \pi_l (o)$ $x_i \leftarrow$ $\epsilon$-$\pi_h(h\_x)$
            \Else
                \State $\theta_i \leftarrow$ $ \pi_l (o)$ $x_i \leftarrow$ $\epsilon$-$\pi_h(h\_x)$
            \EndIf
            \State $action \leftarrow (x_i,\theta_i)$
            \State  $r(t),o,done \leftarrow  env.step(action)$
            \If{$reward>0$}
                \State save $(T_i,o,\theta_i)$ 
            \EndIf
            \If{$done$}
                \State  $data \leftarrow under\_sample(data)$
                \If{$flag<0.5$}    
                    \State  Update $\pi_l $
                \Else
                    \State  Update $\pi_l $, $\pi_h$
                \EndIf
            \EndIf
        \EndWhile  
    \EndWhile  
    \end{algorithmic}
\end{algorithm}

\subsection{Unbalanced Data Distribution Problem}
\par Long-horizon tasks are sequential decision problems. It is divided into some meta-tasks. Only after the first meta-task is completed, the second task can be done. So there is already much data for the first task when the second task data is collected. This problem leads to data imbalance and reduces the stability of the training process. The meta-task that has more data would affect the training of the other tasks.
\par This problem can be resolved by random under-sampling. Random under-sampling is a method that aims to balance class distribution through the random elimination of majority class examples. Once the data of new meta-tasks is collected, the agent will under-sample the data of old meta-tasks.

\subsection{Exploitation Method}
\par The first exploration method uses a traditional method to obtain new positive samples. Low-level policy and high-level policy use the epsilon greedy method to explore simultaneously in one episode. The pseudocode is given in Algorithm \ref{Algorithm 1}. There is a problem when using this method. When the low-level policy is exploring, the high-level policy is also exploring. That means even if the low-level policy is perfect, it still needs to exploit this policy every time to provide a guarantee for the high-level policy to explore the correct sequence of actions successfully. 
\par In order to understand this problem intuitively, We assume that $\epsilon$ is 0.7, which means that the probability of exploration in each episode is 0.7, and the length of the action sequence is 5. Only when the perfect low-level policy is used each time; will the entire task be finished; this probability is 0.00243. Moreover, if we use a small $\epsilon $, the low policy's exploration rate will decrease. So this method performs badly in the long-horizon task. 
\par The exploration method is improved by making the low-level policy and the high-level policy explore separately. If the low-level policy is exploring, the high-level policy uses the greedy method and vice versa. The exploration method is also epsilon greedy. The pseudocode is given in Algorithm \ref{Algorithm 2}.

\label{subsec:a}

\section{EXPERIMENTS}
\par In this section, we shall describe the experimental setup, report the experimental results and analyze the performance of the skill-sequence-dependent policy. The experiments demonstrate that this policy successfully solves a long-horizon task and learns faster than the other baselines. It also shows that the improved exploration method and the under-sampling method are both essential for the policy. 
\begin{figure}[t] %H为当前位置，!htb为忽略美学标准，htbp为浮动图形
\centering %图片居中
\includegraphics[width=0.48\textwidth]{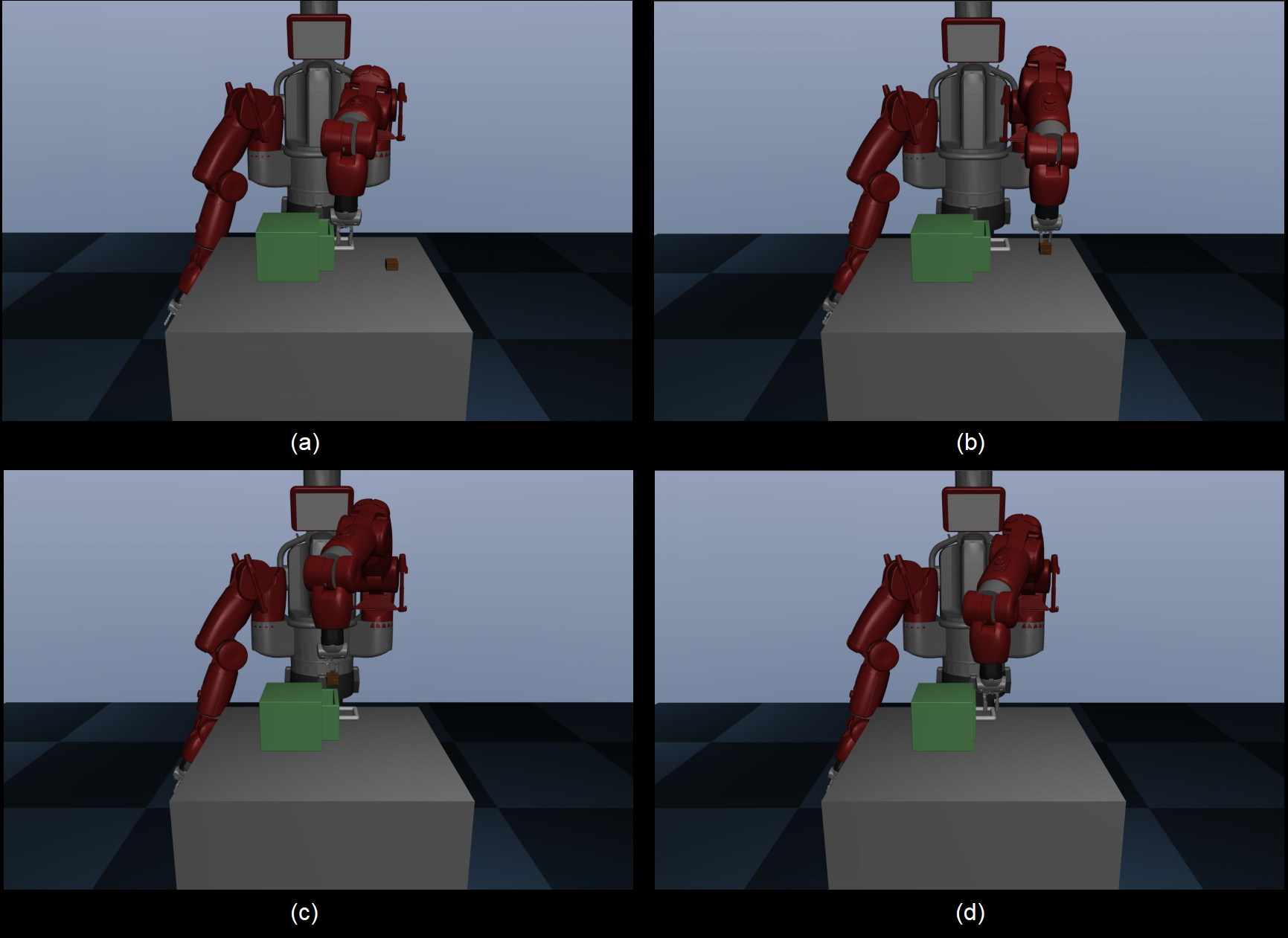} %插入图片，[]中设置图片大小，{}中是图片文件名
\caption{Demonstration of the robot. (a) shows that the robot opens the drawer. (b) shows that the robot grasps the block and lift. (c) shows that the robot put the block into a drawer. (d) shows that the robot finished the long horizon task successfully.} %最终文档中希望显示的图片标题
\label{Fig3} %用于文内引用的标签
\end{figure}

\begin{table}[tbp]
	\centering  % 表居中
	\begin{tabular}{lccc}  % {lccc} 表示各列元素对齐方式，left-l,right-r,center-c
\hline  % 在表格最上方绘制横线
Skill&Details\\
\hline  % 在表格最上方绘制横线
Pull&Pull something from the position to a direction.\\
\hline  %在第一行和第二行之间绘制横线
Grasp & Move the gripper to the position and close it.\\
\hline % 在表格最下方绘制横线
Push & Push something from the position to a direction.\\
\hline % 在表格最下方绘制横线
Put & Move the gripper to the position and open it.\\
\hline % 在表格最下方绘制横线
	\end{tabular}
	\caption{Skills and details of the skills.}
		\label{tab}
\end{table}

\subsection{Experimental Setup}
\par Our approach is evaluated in Mujoco. The implementation of the environment references the Robosuite \cite{zhu2020robosuite} and Meta-World \cite{yu2020meta}, which provide open-source simulated environments for robot learning. The task is putting the block into a drawer and closing the drawer.
\par We define four skills for this task. The details are given in Table \ref{tab}. The parameters of them are $(x,y)$ position in Cartesian coordinates. Because the drawer's direction is fixed, the problem is simplified by fixing the direction of pull and push.

\par The state is the Cartesian coordinates of the block and the drawer. The action is the chosen skill and the parameter for the skill. There are four stages based on these skills. Stage 1: The robot opens the drawer. Stage 2: The robot grasps the block. Stage 3: The robot puts the block into the drawer. Stage 4: The robot closes the drawer.  

The reward function is as follows:
$$
r(t)=
             \begin{cases}
             60 & \text{Stage one is finished} \\  
             70 & \text{Stage two is finished}\\  
             80 & \text{Stage three is finished}\\  
             100 & \text{Stage four is finished}\\
             -2 & \text{Take one invalid step}\\
             -8& \text{Have a collision with something}
             \end{cases}  
$$
\par The biggest reward is 310, which means the robot can finish the task successfully without taking one invalid step.

\begin{figure}[t] %H为当前位置，!htb为忽略美学标准，htbp为浮动图形
\centering %图片居中
\includegraphics[width=0.48\textwidth]{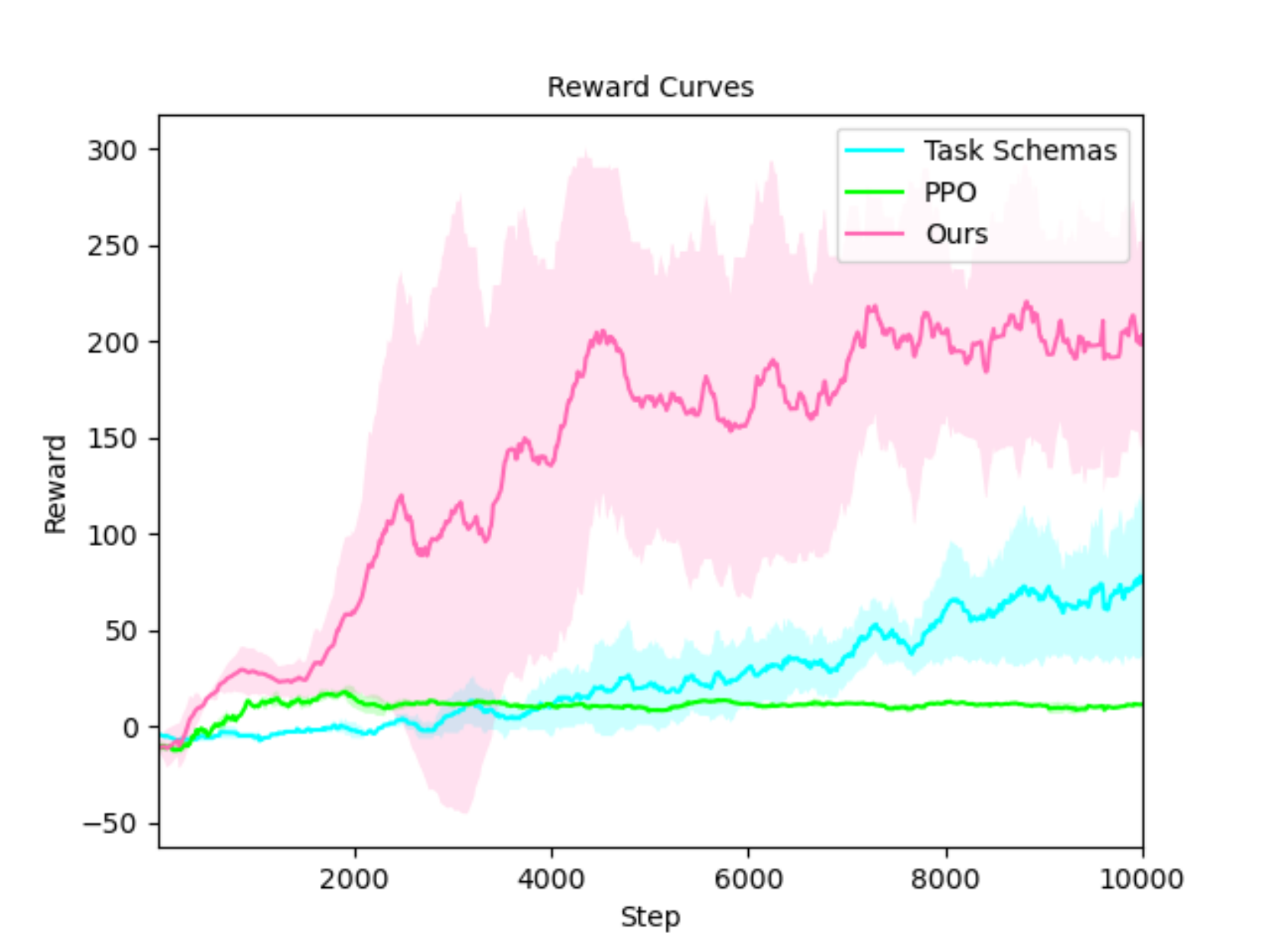} %插入图片，[]中设置图片大小，{}中是图片文件名
\caption{Reward curves for three algorithms. Reward curves can evaluate the number of invalid steps, success rates of four stages, and collision. Our method(pink) is better than the task schema(blue) and PPO(green).} %最终文档中希望显示的图片标题
\label{Fig4} %用于文内引用的标签
\end{figure}

\begin{figure}[t] %H为当前位置，!htb为忽略美学标准，htbp为浮动图形
\centering %图片居中
\includegraphics[width=0.48\textwidth]{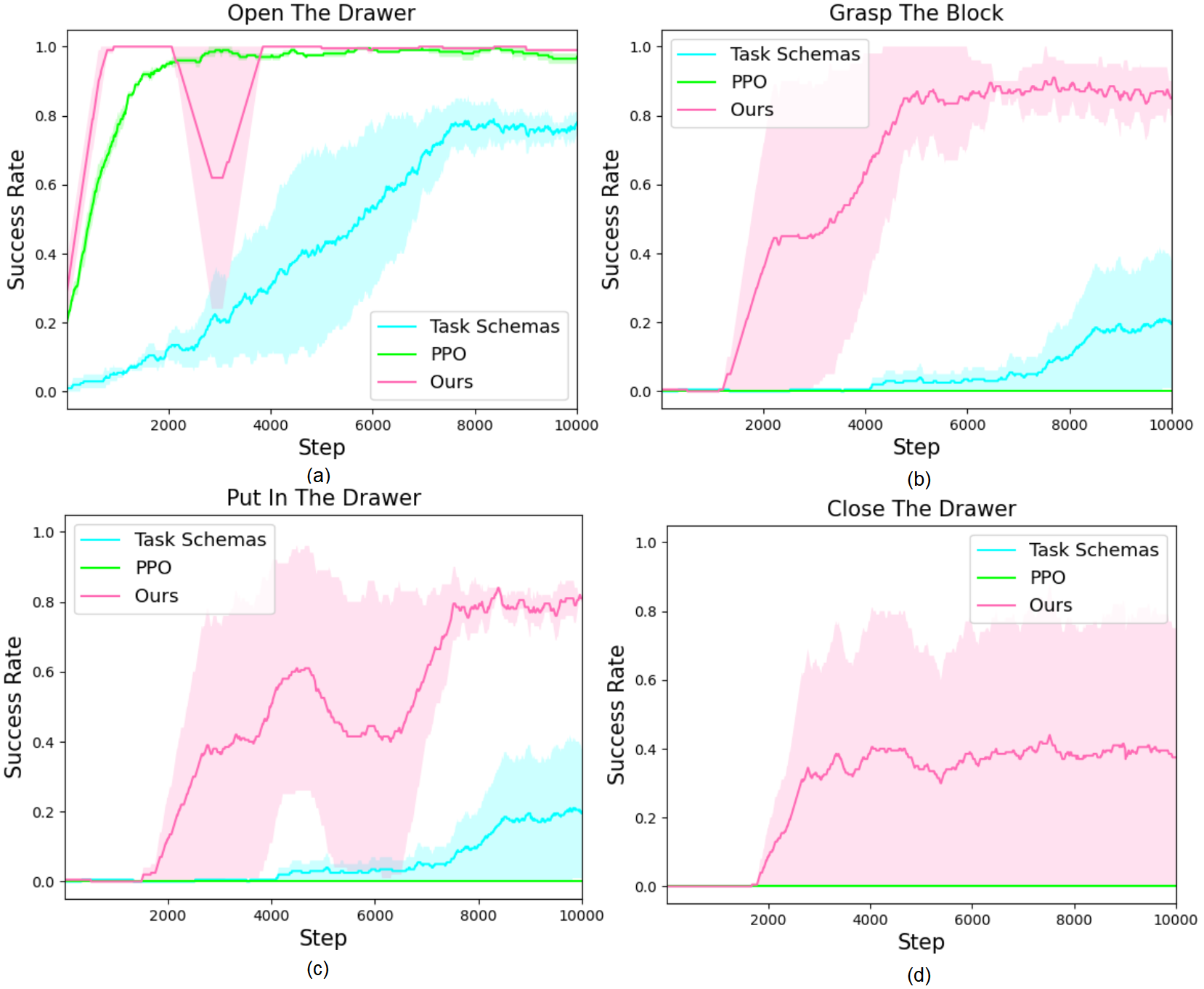} %插入图片，[]中设置图片大小，{}中是图片文件名
\caption{Success rate curves for three algorithms. (a), (b), (c), (d) show the success rate for four stages, respectively. The curve of PPO(green) coincides with the task schema(blue) at the last stage. Therefore only the green one is shown.} %最终文档中希望显示的图片标题
\label{Fig5} %用于文内引用的标签
\end{figure}

\begin{figure}[t] %H为当前位置，!htb为忽略美学标准，htbp为浮动图形
\centering %图片居中
\includegraphics[width=0.48\textwidth]{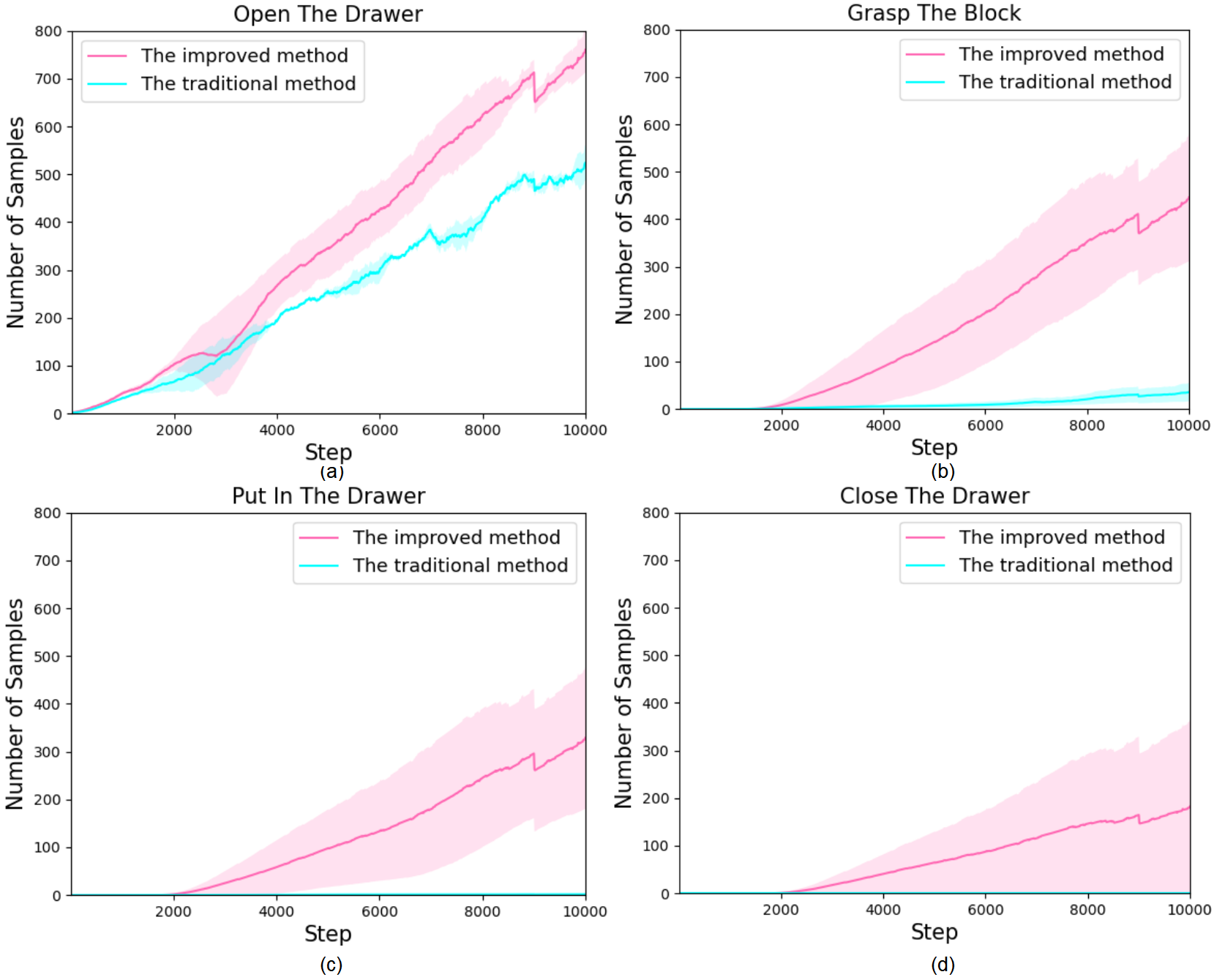} %插入图片，[]中设置图片大小，{}中是图片文件名
\caption{The number of samples shows the efficiency of exploration. } %最终文档中希望显示的图片标题
\label{Fig6} %用于文内引用的标签
\end{figure}

\begin{figure}[t] %H为当前位置，!htb为忽略美学标准，htbp为浮动图形
\centering %图片居中
\includegraphics[width=0.48\textwidth]{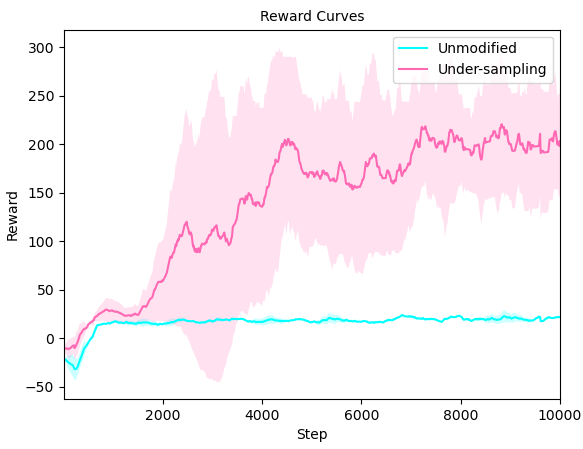} %插入图片，[]中设置图片大小，{}中是图片文件名
\caption{The reward is used  as the evaluating indicator. The under-sampling method(pink) outperforms the unmodified method(blue).} %最终文档中希望显示的图片标题
\label{Fig7} %用于文内引用的标签
\end{figure}

\subsection{Details on Our Policy and Baseline}
\par We train the high-level policy with Q learning and the low-level policy with MLP. The $\epsilon$ for Q learning is 0.15. The MLP contains three hidden layers. The first layer has 32 neurons. The second one has 64 neurons, and the last one also has 32 neurons. The learning rate is 0.0001.
\par The efficiency of our method is compared  to a policy trained using Proximal Policy Optimization(PPO) \cite{schulman2017proximal}. PPO is a family of PG methods for reinforcement learning. It has been shown to work well for continuous control tasks \cite{heess2017emergence}, \cite{melo2019learning}. Our method is also compared to a baseline \cite{chitnis2020efficient} that decomposes the learning process into a state-independent task schema and policy to choose the skills' parameters. Their hierarchical policy finishes many challenging manipulation tasks such as lifting an aluminum tray, opening a glass jar, and picking up a large soccer ball. They use the same MLP structure and reward function as our policy. Their actions are defined as the same as our policy.

\subsection{Experiment Results}
\par \textbf{Comparison of our algorithm and baselines}
\par Our algorithm is evaluated in the long-horizon task that described before. The robot's performance is demonstrated in Figure \ref{Fig3}. It can finish the long horizon task successfully through training. 
\par We compare their reward values and success rates in training. The result is shown in Figure \ref{Fig4} and Figure \ref{Fig5}. It is worth noting that the reward and success rate are related but different since the robot may take many invalid steps but achieve success at last. 
\par Our algorithm significantly improves the efficiency of learning. The reward of our method is consistently higher than the other methods. The success rate is divided into four stages for evaluation.
The success rates of our method and PPO can achieve 1 in the first stage. The success rate of task schema is lower than PPO in the first stage.  However, the performance of task schema is better than PPO in the other three stages. In the last three stages, our method outperforms other baselines with success rate achieving 0.9, 0.8, and 0.4, respectively. It is also the only method that completes the last stage's task. It can ultimately converge in the first three stages. However, it has a high variance in stage four as the characteristic of long-horizon tasks. 

\par \textbf{Sampling efficiency of improved exploration method}
\par The exploration method influences the sample number and learning efficiency. To demonstrate the efficiency of our improved method, we shall compare it with the unmodified method by the number of samples. Both of them are tested in the same simulation environment with the same random seed.
\par The result is given in Figure \ref{Fig6}. The number of samples explored by our improved method is consistently higher than explored by the traditional method in all stages. Our improved method(pink) and the traditional method(blue) have similar efficiency in the first stage. However, the improved method outperforms the traditional method in the last three stages. Although our improved method has a high variance in stage four, it still performs better than the unmodified method. The unmodified method fails to sample in stage four, and the numbers of its samples in stage two and stage three are less than 50. The results indicate that our improved method has significant advantages in a long-horizon manipulation task.

\par \textbf{The effect of under-sampling}
\par There are four stages in this long-horizon task. As the last three stages are more challenging than the first stage, positive samples are sparse in the last three stages. Figure \ref{Fig6} shows this phenomenon. There are already more than one hundred samples for stage one when the agent collects the first sample for stage two. The same phenomenon is shown in stage three and stage four. So this becomes a data imbalance problem. 
\par We use the under-sampling method to balance the data distribution. We compare it with the unmodified method. Both of them use the improved exploration method to prevent the influence of exploration. The result is given in Figure \ref{Fig7}. The reward of the under-sampling method is consistently higher than the unmodified method. The result shows that the under-sampling method is essential for our method.

\subsection{Analysis of the high variance}
\par The shaded part in the figure shows the variance of every method. We find that our method has a higher variance in the reward curves than the other methods. However, this does not mean that our method is more unstable than the other methods. The reasons for the high variance are given in this section.
\par The data in the success rate curves describes the details of every algorithm in all stages. We analyze the success rate curves in the first three stages(Figure \ref{Fig5} (a) (d) (c)). Our method has a high variance in the initial time. However, when the steps achieve 8000, our method becomes stable and has a low variance. Furthermore, the task schema method has a high variance after 8000 steps, and the variance is still increasing. 
\par Subsequently, we analyze the success rate curves in the last stage(Figure \ref{Fig5} (d)). Only our method can successfully finish the task in the last stage. As it is difficult to be finished, our method fails to execute the task sometimes. Therefore, the reward curve for our method has a large variance. The success rates of the other methods are 0 all the time. Therefore, the variances of their success rate are also 0, making their reward curves have a low variance. But in fact, they have bad performance in the long-horizon task as they cannot complete the task in the last stage.

\section{CONCLUSIONS}
\par In this research, a skill-sequence-dependent approach is proposed for robotic long-horizon manipulation and evaluated in simulation with a Baxter robot that put a block into a drawer. The corresponding training process for this policy is also given.  It significantly outperforms the PPO method and the task schema method. We demonstrate that the improved exploration method samples significantly faster than the traditional method, and the under-sample method effectively solves the unstable training problem. 
\par There are still some areas to be improved in the current results. Our manipulation meta-tasks are simple and can be determined by two parameters ($x,y$ in Cartesian coordinates). Moreover, the observation that we use is the ground state, which means the robot needs to obtain the object's position by camera or other sensors. For future work, we aim to focus on contact-rich long-horizon manipulation which has more parameters and finishing it through visual planning and acting.

\bibliographystyle{IEEEtran}
\bibliography{references}

\end{document}